\documentclass{article}

 \usepackage[final]{nips_2018}

\usepackage[utf8]{inputenc} 
\usepackage[T1]{fontenc}    
\usepackage{hyperref}       
\usepackage{url}            
\usepackage{booktabs}       
\usepackage{amsfonts}       
\usepackage{nicefrac}       
\usepackage{microtype}      

\usepackage{xcolor}
\hypersetup{
    colorlinks,
    linkcolor={red!50!black},
    citecolor={blue!50!black},
    urlcolor={blue!80!black}
}

\title{Expanding search in the space of empirical ML}

\author{
  Bronwyn L Woods \\
  Duo Security\\
  \texttt{bwoods@duo.com} \\
}

\begin{document}

\maketitle

\begin{abstract}
  As researchers and practitioners of applied machine learning, we are given a set of requirements on the problem to be solved, the plausibly obtainable data, and the computational resources available. We aim to find (within those bounds) reliably useful combinations of problem, data, and algorithm. An emphasis on algorithmic or technical novelty in ML conference publications leads to exploration of one dimension of this space. Data collection and ML deployment at scale in industry settings offers an environment for exploring the others. Our conferences and reviewing criteria can better support empirical ML by soliciting and incentivizing experimentation and synthesis independent of algorithmic innovation. 
\end{abstract}

\section{Introduction}

Consider a division of the practice of empirical machine learning into three dimensions: 

\begin{description}
\item[Target.] The specification of the problem to be solved, including the scope of the relevant population, assumptions about environment, form of the expected output, and measure of success.
\item[Data.]  \hskip 7pt  An actual dataset instantiated from a particular sampling scheme, measurement methodology, and (frequently) labeling effort.
\item[Algorithm.] The model formalism, hyperparameter selection, and training methodology.
\end{description}

Take any point in the above space and you have a potential machine learning system, though one of questionable expected utility. A goal of applied research is to find optima, but there are many ways to explore the space. Consider just two of many possible representative personas:

\hangindent 6pt \emph{Grace} is a grad student, heavily constrained by publicly available datasets and the corresponding predefined targets. Her work is largely empirical, and she is evaluated based on novel contributions to the field. She demonstrates success by developing new model architectures and comparing their outputs favorably to previously reported methods across several tasks and benchmarks.

\hangindent 6pt \emph{Irene} is an industry data scientist working on ML powered product features. She has access to product data, and some flexibility to alter the target and extracted training dataset. She demonstrates success by designing a useful system that performs as expected for all users. She is incentivized to use established (implemented) algorithms when possible. To avoid failures in the open world production environment, she spends substantial time understanding trends and artifacts in the performance of these algorithms across user populations and datasets.

Arguably, the field of machine learning is the study of the algorithmic dimension of this space. Indeed, the targets and datasets often manifest in conference papers as externally defined challenges, benchmarks, and motivating applications. We might characterize Grace as a researcher and Irene as a practitioner. However, there is a growing empirically motivated and measured branch of ML. As such, we should consider whether our conference discourse and incentives support the full spectrum of relevant empirical research.

\section{Empirical machine learning}

Machine learning has achieved impressive results from methods that are largely empirically motivated and evaluated. This has spawned active discussion on improving empirical methodology in the field. A common theme is the importance of identifying the source and magnitude of gains in reported results (eg \cite{Lipton2018a}, \cite{Sculley2018},  \cite{Rahimi2018}). Conference papers are often presented as \emph{wins} on some benchmark task relative to the "state of the art" as represented by a collection of results from recent papers. The implications of sharing these wins as research are that
\begin{description}
\item[(a)] the difference in performance is related to innovations that are the subject of the paper, and 
\item[(b)] there are likely adjacent or similar tasks for which those innovations would also be helpful. 
\end{description}

Much of the existing discussion on empiricism centers on implication (a) above. Many authors encourage increased emphasis on ablation studies and hyperparameter tuning for benchmarks. \cite{Melis2018} show that with sufficient tuning older LSTM architectures beat newer models, despite previously reported results to the contrary. \cite{Henderson2017} show similar impacts of hyperparameter tuning and other sources of variability in random seeds, compute environments, and codebases for deep reinforcement learning. As the magnitude of the gains over previous results gets smaller, differentiating signal from this kind of experimental noise becomes increasingly critical.

Addressing implication (b), \cite{Lipton2018a} draw attention to studies of the benchmark tasks and datasets themselves. For instance, \cite{Chen2016a} and \cite{Kaushik2018} try to quantify the level of language understanding implied by good performance on popular benchmark tasks for that domain. Similarly, \cite{Torralba2011} quantify bias in image datasets intended to represent the breadth of the visual world.

Many of the suggestions in this ongoing discussion focus on improving the quality, completeness, and accessibility of experimental results within papers that present algorithmic innovations. I support this goal, but also argue that the focus on controlling variability within the training, tuning and computation of models fails to capture the importance of empirical work exploring the \emph{target} and \emph{data} dimensions of ML as defined earlier. In the remainder of this section I attempt to motivate this claim through several observations. 

\subsection{Empirical results depend on datasets}

Individual empirical results depend on datasets. This may seem obvious, but all too often the \emph{target} and \emph{data} dimensions are conflated and called the \emph{problem}. Broad goals like Automated Essay Scoring become synonymous with performance on small collections of benchmark datasets and we easily forget (or at least de-emphasize) that those datasets are themselves only one sample supporting one target within that problem domain.

There are obvious reasons for this pattern. Data is expensive. Most researchers do not have the luxury of sampling new data on demand, let alone paying for labels or annotations. Notions of reproducibility (section \ref{sec:reproducibility}) and benchmark comparisons (section \ref{sec:benchmarks}) are made substantially simpler by widely shared static datasets. 

Nevertheless, we should not forget data derived sources of variability in experimental ML. Especially within smaller datasets, cross-validation splits can have nontrivial impacts on measured performance. More broadly, each dataset as a whole is a finite sample from some population. This sampling introduces variability. We can encourage authors to report variability from subsampling methods, and justify widespread adoption of benchmarks as a means of controlling for sampling variation.

But variability isn't bad. If we are learning about our algorithms empirically, testing under realistic variability is essential to discovering generalizable patterns. The benchmark literature gives us one picture of the landscape of algorithm performance. Holding the algorithms constant, how does that picture change under small perturbations to the data (eg resampling from the same population) or the target (eg asking for the same predictions on data from a different population)? What stable patterns in both data and algorithm space does this reveal? This kind of experimentation is hard or impossible to do when constrained to public benchmark datasets. The researchers with the resources and learned intuition for this work may not be the same as those doing innovative algorithmic development.

\subsection{Benchmarks are a double-edged sword}
\label{sec:benchmarks}

Public, shared benchmark datasets are hugely important to the field of machine learning. By fixing the input data and the prediction target, we remove a large source of variability (discussed above) and allow disparate researchers to quantitatively compare results. 

Benchmarks create competitions, and competitions are compelling. But as eloquently stated in \cite{Sculley2018}, \emph{"a moment of reflection recalls that the goal of science is not wins, but knowledge".} Year over year, we incrementally push the dial upwards on benchmark metrics. How do we ensure that by doing so we are still creating knowledge?

Despite an understanding of overfitting on the scale of individual models, we have very few safeguards in place to detect or correct for overfitting to our benchmarks as a field. In data rich environments with relatively few algorithms of interest (such as many companies), benchmarks can be cycled periodically and metrics recomputed. In the context of the broader research community, new datasets are difficult or impossible to release publicly (often for legal reasons), and there is no convenient mechanism for recomputing results across existing papers. In practice, individual benchmark datasets are persistent and hugely influential.

Without benchmarks we are reduced to rhetorical argument and incomparable anecdotes. With benchmarks, we have quantitative measures that, without care, give implicit permission to skip the work of measuring anything outside of the closed world of a single dataset.

\subsection{Reproducibility is a group effort}
\label{sec:reproducibility}

The idea of reproducibility in ML has long been both valued and controversial. The terminology is confused (\cite{Plesser2018}), but an important contrast is that between the ability to exactly recreate the results of a particular paper (I will call this repeatability) and the ability to reach the same conclusions despite variation in the components of the experiment assumed to be irrelevant (I will call this reproducibility). This distinction is discussed at length in \cite{Drummond2009}, who takes the relatively extreme position that repeatability has little value.

In contrast, a lot of recent focus has been placed on repeatability. \cite{Tatman2018} outline a taxonomy for repeatable (\emph{reproducible}, in their terminology) ML, with the lowest level being careful specification of the algorithmic and experimental conditions and the highest level being the release of code and data together with a compute environment. Machine learning is unlike many experimental sciences in that such an extreme level of repeatability is possible. We shouldn't ignore this ability. Recent advances in "reproducible reporting" and data and code sharing help find errors, aid in building understanding of new approaches, and make ablation studies of algorithmic components substantially easier. But this comes at a cost.

One cost of a focus on repeatability is that we may take focus away from reproducibility. Repeatability removes variability; reproducibility requires it. In a sense, we get the best measure of reproducibility by using the channel that \cite{Tatman2018} defined as least repeatable: an un-runnable specification of algorithm, data properties, and necessary peripheral settings. If those specifications are insufficient for independent research to reach the same conclusion, we are missing something in our understanding of the phenomenon. Outside of ML, a commonly cited example of such a confounder is the \cite{Sorge2014} finding that the sex of the human experimenter conducting studies of rats had substantial effects on their stress response. This was a factor assumed to be irrelevant that was in fact hindering the reproducibility of results. 

As highlighted in \cite{Irpan2018}, requirements of demonstrated repeatability through careful release of artifacts and/or reproducibility through extensive experimentation under different conditions also have a cost in terms of burden on individual researchers. Irpan argues that this cost is sufficiently large that we should not require reproducibility demonstrations or guarantees from all papers. A different question is whether we should require it of ourselves as a field. 

A useful reproducible result in ML is a statement that applies to a non-point-mass region of the the [target $\times$ data $\times$ algorithm] space. We can hold the data constant to understand the algorithm dimension; we should also hold the algorithm constant to understand performance across the data dimension. But that understanding need not derive from a single paper or a single group or researcher. The computational or data resources that are overly burdensome to Grace the grad student may be mundane to Irene in industry.

\section{Recommendations}

Empirical results are central to ML. Though extremely useful, benchmark datasets have a huge influence over the course of ML research. Open-world exploration of algorithm performance through continuous or repeated data collection is a relatively uncommon ability. We value reproducibility, but cannot practically impose its full cost on individual researchers as a prerequisite to publication. How can we adapt our existing conference and incentive structure accordingly? I give a few suggestions, and pointers to where we can expand existing efforts.

\subsection{(Re)define novelty}

ML conferences heavily prioritize "novel" research. The word shows up in nearly every call for papers (CFP), though it is rarely if ever precisely defined. I think it is most frequently taken to mean a requirement for algorithmic or model novelty. This can be discouraging to researchers whose contributions would be of different experimental forms, such as drawing inferences from the behavior of known algorithms across data collected from multiple populations or with different measurement methodologies. 
An example of reducing the emphasis on algorithmic novelty is the KDD Applied Data Science track. The CFP\footnote{https://www.kdd.org/kdd2018/calls/view/kdd-2018-call-for-applied-data-science-papers} includes an explicit category for observational papers which "describe important insights into the input data and/or the performance" of ML systems. But even that track also states that a "base requirement is that the paper  describes a novel artifact that has real users", again discouraging knowledge created from synthesis and experimentation on existing methods and systems.

\subsection{Make space for synthesis}

Traditional ML conference formats and reviewing criteria are not well matched to work lacking in algorithmic novelty but rich in knowledge created from comparisons of existing algorithms. However it is exactly by incentivizing this sort of work that we can take some of the burden of reproducibility and expensive experimental rigor off of individual submissions exploring the algorithmic space. 

An example of incentivizing the work of synthesis is the Systematization of Knowledge (SoK) track at the IEEE Symposium on Security and Privacy. As described in the CFP\footnote{https://www.ieee-security.org/TC/SP2018/cfpapers.html}, the track selects papers "that provide an important new viewpoint on an established, major research area, support or challenge long-held beliefs in such an area with compelling evidence, or present a convincing, comprehensive new taxonomy of such an area." The non-traditional ML publication Distill\footnote{https://distill.pub/about/} has a similar aim.


\subsection{Incentivize publishing open world experimentation}

Conferences could solicit open world experimental results, explicitly complementing innovation in the algorithmic space with observation and experimentation over data and target dimensions. An important consideration here is the position of the researchers likely to have such results to offer and the incentives that can apply. 

My experience is that this is the area where industry researchers, both in and outside the large corporate research labs, have much to contribute. For many of these data scientists, machine learning engineers, and research scientists, conference publication itself has a relatively small impact on their professional reputation and success. But they have much to gain from guiding the comparatively vast academic research community toward the practical gaps in existing algorithms. As we innovate on ways to handle the distribution of the increasing volume of new algorithmic work (\cite{Soergel2013, Lee2018}), conferences have the opportunity to facilitate collaboration between open world experimenters providing a picture of reproducibility and important algorithmic gaps, and algorithmic researchers innovating to address those needs.

Good open world experimentation is made easier, but not replaced, by better ML toolkits and shared algorithmic code. The results of this work are strengthened, but not replaced, by the creation of new or more representative shared datasets. Strong empirical ML requires contributions from researchers with different resources, incentives, and domains of learned intuition and expertise.

%

%
%
%
%
%
%
%
%
\small
\bibliographystyle{plainnat}
\bibliography{/Users/bwoods/Dropbox/papers/bibtex/Duo-ML_trends}

\end{document}